\documentclass[sigconf]{acmart}

\usepackage{booktabs} 
\usepackage{comment}
\usepackage{bm}
\usepackage{amsthm,amsmath,amsfonts,dsfont}
\usepackage{balance}

\begin{document}
\emergencystretch 1em

\title{CARLS: \underline{C}ross-platform \underline{A}synchronous \underline{R}epresentation \underline{L}earning \underline{S}ystem}

\author{Chun-Ta Lu$^*$, Yun Zeng$^*$, Da-Cheng Juan$^*$, Yicheng Fan, Zhe Li, Jan Dlabal, Yi-Ting Chen, Arjun Gopalan, Allan Heydon, Chun-Sung Ferng, Reah Miyara, Ariel Fuxman, Futang Peng, Zhen Li, Tom Duerig, Andrew Tomkins}
\affiliation{%
  \institution{Google Research}
  \city{Mountain View}
  \state{CA}
}
\email{{chunta,xzeng,dacheng,yichengfan,lzhe,dlabal,yitingchen,arjung,aheydon,csferng,reahm,afuxman,futangpeng,zhenli,tduerig,tomkins}@google.com}

\renewcommand{\shortauthors}{}

\begin{abstract}
\let\thefootnote\relax\footnotetext{* These authors contributed equally.}
Innovation on learning systems has been a key propellant in advancing and scaling up state-of-the-art neural models.
In this work, we propose CARLS, a novel framework for augmenting the capacity of existing deep learning frameworks by enabling multiple components---model trainers, knowledge makers and knowledge banks---to concertedly work together in an asynchronous fashion across hardware platforms. 
The proposed CARLS is particularly suitable for learning paradigms where model training benefits from additional knowledge inferred or discovered during training, such as node embeddings for graph neural networks or reliable pseudo labels from model predictions. 
We also describe three learning paradigms---semi-supervised learning, curriculum learning and multimodal learning---as examples that can be scaled up efficiently by CARLS. 
One version of CARLS has been open-sourced and available for download at: \\
\href{https://github.com/tensorflow/neural-structured-learning/tree/master/research/carls}{\small \color{blue}{github.com/tensorflow/neural-structured-learning/tree/master/research/carls}}. 

\end{abstract}

\maketitle


\newcommand\reminder[1]{\textcolor{blue}{#1}}
\newcommand{\ie}{{\textit i.e.}}
\newcommand{\eg}{{\textit e.g.}}

\newcommand{\ourmodel}[0]{CARLS}


\newcommand{\figleft}{{\em (Left)}}
\newcommand{\figcenter}{{\em (Center)}}
\newcommand{\figright}{{\em (Right)}}
\newcommand{\figtop}{{\em (Top)}}
\newcommand{\figbottom}{{\em (Bottom)}}
\newcommand{\captiona}{{\em (a)}}
\newcommand{\captionb}{{\em (b)}}
\newcommand{\captionc}{{\em (c)}}
\newcommand{\captiond}{{\em (d)}}

\newcommand{\newterm}[1]{{\bf #1}}

\def\figref#1{figure~\ref{#1}}
\def\Figref#1{Figure~\ref{#1}}
\def\twofigref#1#2{figures \ref{#1} and \ref{#2}}
\def\quadfigref#1#2#3#4{figures \ref{#1}, \ref{#2}, \ref{#3} and \ref{#4}}
\def\secref#1{section~\ref{#1}}
\def\Secref#1{Section~\ref{#1}}
\def\twosecrefs#1#2{sections \ref{#1} and \ref{#2}}
\def\secrefs#1#2#3{sections \ref{#1}, \ref{#2} and \ref{#3}}
\def\eqref#1{equation~\ref{#1}}
\def\Eqref#1{Equation~\ref{#1}}
\def\plaineqref#1{\ref{#1}}
\def\chapref#1{chapter~\ref{#1}}
\def\Chapref#1{Chapter~\ref{#1}}
\def\rangechapref#1#2{chapters\ref{#1}--\ref{#2}}
\def\algref#1{algorithm~\ref{#1}}
\def\Algref#1{Algorithm~\ref{#1}}
\def\twoalgref#1#2{algorithms \ref{#1} and \ref{#2}}
\def\Twoalgref#1#2{Algorithms \ref{#1} and \ref{#2}}
\def\partref#1{part~\ref{#1}}
\def\Partref#1{Part~\ref{#1}}
\def\twopartref#1#2{parts \ref{#1} and \ref{#2}}

\def\ceil#1{\lceil #1 \rceil}
\def\floor#1{\lfloor #1 \rfloor}
\def\1{\bm{1}}
\newcommand{\train}{\mathcal{D}}
\newcommand{\valid}{\mathcal{D_{\mathrm{valid}}}}
\newcommand{\test}{\mathcal{D_{\mathrm{test}}}}

\def\eps{{\epsilon}}

\def\reta{{\textnormal{$\eta$}}}
\def\ra{{\textnormal{a}}}
\def\rb{{\textnormal{b}}}
\def\rc{{\textnormal{c}}}
\def\rd{{\textnormal{d}}}
\def\re{{\textnormal{e}}}
\def\rf{{\textnormal{f}}}
\def\rg{{\textnormal{g}}}
\def\rh{{\textnormal{h}}}
\def\ri{{\textnormal{i}}}
\def\rj{{\textnormal{j}}}
\def\rk{{\textnormal{k}}}
\def\rl{{\textnormal{l}}}
\def\rn{{\textnormal{n}}}
\def\ro{{\textnormal{o}}}
\def\rp{{\textnormal{p}}}
\def\rq{{\textnormal{q}}}
\def\rr{{\textnormal{r}}}
\def\rs{{\textnormal{s}}}
\def\rt{{\textnormal{t}}}
\def\ru{{\textnormal{u}}}
\def\rv{{\textnormal{v}}}
\def\rw{{\textnormal{w}}}
\def\rx{{\textnormal{x}}}
\def\ry{{\textnormal{y}}}
\def\rz{{\textnormal{z}}}

\def\rvepsilon{{\mathbf{\epsilon}}}
\def\rvtheta{{\mathbf{\theta}}}
\def\rva{{\mathbf{a}}}
\def\rvb{{\mathbf{b}}}
\def\rvc{{\mathbf{c}}}
\def\rvd{{\mathbf{d}}}
\def\rve{{\mathbf{e}}}
\def\rvf{{\mathbf{f}}}
\def\rvg{{\mathbf{g}}}
\def\rvh{{\mathbf{h}}}
\def\rvu{{\mathbf{i}}}
\def\rvj{{\mathbf{j}}}
\def\rvk{{\mathbf{k}}}
\def\rvl{{\mathbf{l}}}
\def\rvm{{\mathbf{m}}}
\def\rvn{{\mathbf{n}}}
\def\rvo{{\mathbf{o}}}
\def\rvp{{\mathbf{p}}}
\def\rvq{{\mathbf{q}}}
\def\rvr{{\mathbf{r}}}
\def\rvs{{\mathbf{s}}}
\def\rvt{{\mathbf{t}}}
\def\rvu{{\mathbf{u}}}
\def\rvv{{\mathbf{v}}}
\def\rvw{{\mathbf{w}}}
\def\rvx{{\mathbf{x}}}
\def\rvy{{\mathbf{y}}}
\def\rvz{{\mathbf{z}}}

\def\erva{{\textnormal{a}}}
\def\ervb{{\textnormal{b}}}
\def\ervc{{\textnormal{c}}}
\def\ervd{{\textnormal{d}}}
\def\erve{{\textnormal{e}}}
\def\ervf{{\textnormal{f}}}
\def\ervg{{\textnormal{g}}}
\def\ervh{{\textnormal{h}}}
\def\ervi{{\textnormal{i}}}
\def\ervj{{\textnormal{j}}}
\def\ervk{{\textnormal{k}}}
\def\ervl{{\textnormal{l}}}
\def\ervm{{\textnormal{m}}}
\def\ervn{{\textnormal{n}}}
\def\ervo{{\textnormal{o}}}
\def\ervp{{\textnormal{p}}}
\def\ervq{{\textnormal{q}}}
\def\ervr{{\textnormal{r}}}
\def\ervs{{\textnormal{s}}}
\def\ervt{{\textnormal{t}}}
\def\ervu{{\textnormal{u}}}
\def\ervv{{\textnormal{v}}}
\def\ervw{{\textnormal{w}}}
\def\ervx{{\textnormal{x}}}
\def\ervy{{\textnormal{y}}}
\def\ervz{{\textnormal{z}}}

\def\rmA{{\mathbf{A}}}
\def\rmB{{\mathbf{B}}}
\def\rmC{{\mathbf{C}}}
\def\rmD{{\mathbf{D}}}
\def\rmE{{\mathbf{E}}}
\def\rmF{{\mathbf{F}}}
\def\rmG{{\mathbf{G}}}
\def\rmH{{\mathbf{H}}}
\def\rmI{{\mathbf{I}}}
\def\rmJ{{\mathbf{J}}}
\def\rmK{{\mathbf{K}}}
\def\rmL{{\mathbf{L}}}
\def\rmM{{\mathbf{M}}}
\def\rmN{{\mathbf{N}}}
\def\rmO{{\mathbf{O}}}
\def\rmP{{\mathbf{P}}}
\def\rmQ{{\mathbf{Q}}}
\def\rmR{{\mathbf{R}}}
\def\rmS{{\mathbf{S}}}
\def\rmT{{\mathbf{T}}}
\def\rmU{{\mathbf{U}}}
\def\rmV{{\mathbf{V}}}
\def\rmW{{\mathbf{W}}}
\def\rmX{{\mathbf{X}}}
\def\rmY{{\mathbf{Y}}}
\def\rmZ{{\mathbf{Z}}}

\def\ermA{{\textnormal{A}}}
\def\ermB{{\textnormal{B}}}
\def\ermC{{\textnormal{C}}}
\def\ermD{{\textnormal{D}}}
\def\ermE{{\textnormal{E}}}
\def\ermF{{\textnormal{F}}}
\def\ermG{{\textnormal{G}}}
\def\ermH{{\textnormal{H}}}
\def\ermI{{\textnormal{I}}}
\def\ermJ{{\textnormal{J}}}
\def\ermK{{\textnormal{K}}}
\def\ermL{{\textnormal{L}}}
\def\ermM{{\textnormal{M}}}
\def\ermN{{\textnormal{N}}}
\def\ermO{{\textnormal{O}}}
\def\ermP{{\textnormal{P}}}
\def\ermQ{{\textnormal{Q}}}
\def\ermR{{\textnormal{R}}}
\def\ermS{{\textnormal{S}}}
\def\ermT{{\textnormal{T}}}
\def\ermU{{\textnormal{U}}}
\def\ermV{{\textnormal{V}}}
\def\ermW{{\textnormal{W}}}
\def\ermX{{\textnormal{X}}}
\def\ermY{{\textnormal{Y}}}
\def\ermZ{{\textnormal{Z}}}

\def\vzero{{\bm{0}}}
\def\vone{{\bm{1}}}
\def\vmu{{\bm{\mu}}}
\def\vtheta{{\bm{\theta}}}
\def\va{{\bm{a}}}
\def\vb{{\bm{b}}}
\def\vc{{\bm{c}}}
\def\vd{{\bm{d}}}
\def\ve{{\bm{e}}}
\def\vf{{\bm{f}}}
\def\vg{{\bm{g}}}
\def\vh{{\bm{h}}}
\def\vi{{\bm{i}}}
\def\vj{{\bm{j}}}
\def\vk{{\bm{k}}}
\def\vl{{\bm{l}}}
\def\vm{{\bm{m}}}
\def\vn{{\bm{n}}}
\def\vo{{\bm{o}}}
\def\vp{{\bm{p}}}
\def\vq{{\bm{q}}}
\def\vr{{\bm{r}}}
\def\vs{{\bm{s}}}
\def\vt{{\bm{t}}}
\def\vu{{\bm{u}}}
\def\vv{{\bm{v}}}
\def\vw{{\bm{w}}}
\def\vx{{\bm{x}}}
\def\vy{{\bm{y}}}
\def\vz{{\bm{z}}}
\def\vtheta{{\bm{\theta}}}

\def\evalpha{{\alpha}}
\def\evbeta{{\beta}}
\def\evepsilon{{\epsilon}}
\def\evlambda{{\lambda}}
\def\evomega{{\omega}}
\def\evmu{{\mu}}
\def\evpsi{{\psi}}
\def\evsigma{{\sigma}}
\def\evtheta{{\theta}}
\def\eva{{a}}
\def\evb{{b}}
\def\evc{{c}}
\def\evd{{d}}
\def\eve{{e}}
\def\evf{{f}}
\def\evg{{g}}
\def\evh{{h}}
\def\evi{{i}}
\def\evj{{j}}
\def\evk{{k}}
\def\evl{{l}}
\def\evm{{m}}
\def\evn{{n}}
\def\evo{{o}}
\def\evp{{p}}
\def\evq{{q}}
\def\evr{{r}}
\def\evs{{s}}
\def\evt{{t}}
\def\evu{{u}}
\def\evv{{v}}
\def\evw{{w}}
\def\evx{{x}}
\def\evy{{y}}
\def\evz{{z}}

\def\mA{{\bm{A}}}
\def\mB{{\bm{B}}}
\def\mC{{\bm{C}}}
\def\mD{{\bm{D}}}
\def\mE{{\bm{E}}}
\def\mF{{\bm{F}}}
\def\mG{{\bm{G}}}
\def\mH{{\bm{H}}}
\def\mI{{\bm{I}}}
\def\mJ{{\bm{J}}}
\def\mK{{\bm{K}}}
\def\mL{{\bm{L}}}
\def\mM{{\bm{M}}}
\def\mN{{\bm{N}}}
\def\mO{{\bm{O}}}
\def\mP{{\bm{P}}}
\def\mQ{{\bm{Q}}}
\def\mR{{\bm{R}}}
\def\mS{{\bm{S}}}
\def\mT{{\bm{T}}}
\def\mU{{\bm{U}}}
\def\mV{{\bm{V}}}
\def\mW{{\bm{W}}}
\def\mX{{\bm{X}}}
\def\mY{{\bm{Y}}}
\def\mZ{{\bm{Z}}}
\def\mBeta{{\bm{\beta}}}
\def\mPhi{{\bm{\Phi}}}
\def\mLambda{{\bm{\Lambda}}}
\def\mSigma{{\bm{\Sigma}}}

\newcommand{\tens}[1]{\bm{\mathsfit{#1}}}
\def\tA{{\tens{A}}}
\def\tB{{\tens{B}}}
\def\tC{{\tens{C}}}
\def\tD{{\tens{D}}}
\def\tE{{\tens{E}}}
\def\tF{{\tens{F}}}
\def\tG{{\tens{G}}}
\def\tH{{\tens{H}}}
\def\tI{{\tens{I}}}
\def\tJ{{\tens{J}}}
\def\tK{{\tens{K}}}
\def\tL{{\tens{L}}}
\def\tM{{\tens{M}}}
\def\tN{{\tens{N}}}
\def\tO{{\tens{O}}}
\def\tP{{\tens{P}}}
\def\tQ{{\tens{Q}}}
\def\tR{{\tens{R}}}
\def\tS{{\tens{S}}}
\def\tT{{\tens{T}}}
\def\tU{{\tens{U}}}
\def\tV{{\tens{V}}}
\def\tW{{\tens{W}}}
\def\tX{{\tens{X}}}
\def\tY{{\tens{Y}}}
\def\tZ{{\tens{Z}}}

\def\gA{{\mathcal{A}}}
\def\gB{{\mathcal{B}}}
\def\gC{{\mathcal{C}}}
\def\gD{{\mathcal{D}}}
\def\gE{{\mathcal{E}}}
\def\gF{{\mathcal{F}}}
\def\gG{{\mathcal{G}}}
\def\gH{{\mathcal{H}}}
\def\gI{{\mathcal{I}}}
\def\gJ{{\mathcal{J}}}
\def\gK{{\mathcal{K}}}
\def\gL{{\mathcal{L}}}
\def\gM{{\mathcal{M}}}
\def\gN{{\mathcal{N}}}
\def\gO{{\mathcal{O}}}
\def\gP{{\mathcal{P}}}
\def\gQ{{\mathcal{Q}}}
\def\gR{{\mathcal{R}}}
\def\gS{{\mathcal{S}}}
\def\gT{{\mathcal{T}}}
\def\gU{{\mathcal{U}}}
\def\gV{{\mathcal{V}}}
\def\gW{{\mathcal{W}}}
\def\gX{{\mathcal{X}}}
\def\gY{{\mathcal{Y}}}
\def\gZ{{\mathcal{Z}}}

\def\sA{{\mathbb{A}}}
\def\sB{{\mathbb{B}}}
\def\sC{{\mathbb{C}}}
\def\sD{{\mathbb{D}}}
\def\sF{{\mathbb{F}}}
\def\sG{{\mathbb{G}}}
\def\sH{{\mathbb{H}}}
\def\sI{{\mathbb{I}}}
\def\sJ{{\mathbb{J}}}
\def\sK{{\mathbb{K}}}
\def\sL{{\mathbb{L}}}
\def\sM{{\mathbb{M}}}
\def\sN{{\mathbb{N}}}
\def\sO{{\mathbb{O}}}
\def\sP{{\mathbb{P}}}
\def\sQ{{\mathbb{Q}}}
\def\sR{{\mathbb{R}}}
\def\sS{{\mathbb{S}}}
\def\sT{{\mathbb{T}}}
\def\sU{{\mathbb{U}}}
\def\sV{{\mathbb{V}}}
\def\sW{{\mathbb{W}}}
\def\sX{{\mathbb{X}}}
\def\sY{{\mathbb{Y}}}
\def\sZ{{\mathbb{Z}}}

\def\emLambda{{\Lambda}}
\def\emA{{A}}
\def\emB{{B}}
\def\emC{{C}}
\def\emD{{D}}
\def\emE{{E}}
\def\emF{{F}}
\def\emG{{G}}
\def\emH{{H}}
\def\emI{{I}}
\def\emJ{{J}}
\def\emK{{K}}
\def\emL{{L}}
\def\emM{{M}}
\def\emN{{N}}
\def\emO{{O}}
\def\emP{{P}}
\def\emQ{{Q}}
\def\emR{{R}}
\def\emS{{S}}
\def\emT{{T}}
\def\emU{{U}}
\def\emV{{V}}
\def\emW{{W}}
\def\emX{{X}}
\def\emY{{Y}}
\def\emZ{{Z}}
\def\emSigma{{\Sigma}}

\newcommand{\etens}[1]{\mathsfit{#1}}
\def\etLambda{{\etens{\Lambda}}}
\def\etA{{\etens{A}}}
\def\etB{{\etens{B}}}
\def\etC{{\etens{C}}}
\def\etD{{\etens{D}}}
\def\etE{{\etens{E}}}
\def\etF{{\etens{F}}}
\def\etG{{\etens{G}}}
\def\etH{{\etens{H}}}
\def\etI{{\etens{I}}}
\def\etJ{{\etens{J}}}
\def\etK{{\etens{K}}}
\def\etL{{\etens{L}}}
\def\etM{{\etens{M}}}
\def\etN{{\etens{N}}}
\def\etO{{\etens{O}}}
\def\etP{{\etens{P}}}
\def\etQ{{\etens{Q}}}
\def\etR{{\etens{R}}}
\def\etS{{\etens{S}}}
\def\etT{{\etens{T}}}
\def\etU{{\etens{U}}}
\def\etV{{\etens{V}}}
\def\etW{{\etens{W}}}
\def\etX{{\etens{X}}}
\def\etY{{\etens{Y}}}
\def\etZ{{\etens{Z}}}

\newcommand{\pdata}{p_{\rm{data}}}
\newcommand{\ptrain}{\hat{p}_{\rm{data}}}
\newcommand{\Ptrain}{\hat{P}_{\rm{data}}}
\newcommand{\pmodel}{p_{\rm{model}}}
\newcommand{\Pmodel}{P_{\rm{model}}}
\newcommand{\ptildemodel}{\tilde{p}_{\rm{model}}}
\newcommand{\pencode}{p_{\rm{encoder}}}
\newcommand{\pdecode}{p_{\rm{decoder}}}
\newcommand{\precons}{p_{\rm{reconstruct}}}

\newcommand{\laplace}{\mathrm{Laplace}} 

\newcommand{\E}{\mathbb{E}}
\newcommand{\Ls}{\mathcal{L}}
\newcommand{\R}{\mathbb{R}}
\newcommand{\emp}{\tilde{p}}
\newcommand{\lr}{\alpha}
\newcommand{\reg}{\lambda}
\newcommand{\rect}{\mathrm{rectifier}}
\newcommand{\softmax}{\mathrm{softmax}}
\newcommand{\sigmoid}{\sigma}
\newcommand{\softplus}{\zeta}
\newcommand{\KL}{D_{\mathrm{KL}}}
\newcommand{\Var}{\mathrm{Var}}
\newcommand{\standarderror}{\mathrm{SE}}
\newcommand{\Cov}{\mathrm{Cov}}
\newcommand{\normlzero}{L^0}
\newcommand{\normlone}{L^1}
\newcommand{\normltwo}{L^2}
\newcommand{\normlp}{L^p}
\newcommand{\normmax}{L^\infty}

\section{Introduction}
\label{sec:introduction}

During the past decade, several deep learning frameworks ~\cite{collobert2002torch,abadi2016tensorflow,pytorch} have been made available to public and accelerating the advancement of many research domains in deep learning~\cite{goodfellow2016deep,deep_learning_nature,deep_learning_in_health_care}.
Despite their successes, cutting-edge technology and emerging applications are always pushing the limits of training capacity supported by the existing frameworks.
For example, the largest T5 model~\cite{2020t5} with 11 billion parameters has achieved state-of-the-art performance on multiple NLP benchmarks, and a near-human score on the SuperGLUE natural language understanding benchmark. 
GPT-3~\cite{gpt3}, an auto-regressive language model with 175 billion parameters, achieves strong performance on many NLP benchmarks without any gradient updates or fine-tuning.
Most recently, the switch transformer~\cite{large_model_switch_transformers} advanced the size of the T5 model to $7\times$ times, leading to a giant model speaking $102$ different languages.

While most of the above efforts focus on training one monolithic model, human brains often solve a cognition problem using multiple pathways with multi-sensory inputs and multiple models~\cite{kandel2000principles}.
Such a multimodal framework has already shown its potential in boosting the performance over the existing monolithic models (\eg, ~\cite{lu202012,radford2learning,jia2021scaling}) and in solving novel problems~\cite{graves2016hybrid}. For example, for vision-language related problems such as visual question answering~\cite{antol2015vqa}, caption-based image retrieval~\cite{young2014image}, and visual
common sense reasoning~\cite{zellers2019recognition}, the vision and language modalities are encoded by its own designated encoder and the two encoders are co-trained jointly. However, implementing these multimodal learning tasks at scale imposes unique challenges, since the computational complexity usually grows with the number of encoders trained, and the advancement of hardware alone cannot keep up with the fast growth of model (or data) size and complexity.
Such a scaling issue manifests itself in other machine learning paradigms such as graph-based neural networks~\cite{kipf2016semi,velivckovic2017graph,bui2018neural}, where each node in a graph is processed by a kernel that can be a simple multilayer perceptron or a complex attention model like BERT~\cite{devlin2018bert}.
Furthermore, the computational cost grows linearly with the size of neighborhoods if no sampling technique is used.

\begin{figure}[t]
\centering
      \includegraphics[width=\linewidth]{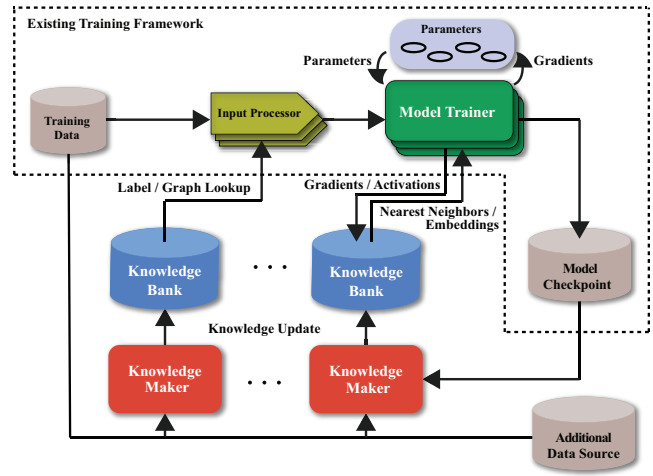}
  \caption{System Overview. CARLS employs three major components: Model Trainer, Knowledge Maker and Knowledge Bank. Models Trainers are the main jobs training and updating model parameters, and Knowledge Makers are running in parallel to compute auxiliary information required for gradient calculation and store the information in Knowledge Bank. Note that these components can be deployed on different platforms (GPUs, TPUs, etc) and communicate with each other in an asynchronous fashion.}
 \label{fig:carls_overview}
\end{figure}

In this work, we propose \emph{CARLS}: a \textbf{C}ross-platform, \textbf{A}synchronous \textbf{R}epresentation \textbf{L}earning \textbf{S}ystem that addresses the challenges in  scaling up these ``giant'' models (\eg, multimodal learning or graph neural networks). 
CARLS is able to (a) accommodate training of very large models with extremely high computational complexity---\eg, graph neural networks with a graph of more than ten billion nodes, each node encoded by a BERT kernel---and (b) accelerate the training by shifting partial workload of inferring information required for gradient calculation from model trainers to ``Knowledge Makers'' deployed in CARLS.

CARLS contains three major components that operate in parallel: \emph{Model Trainer}, \emph{Knowledge Maker} and \emph{Knowledge Bank}.
Figure~\ref{fig:carls_overview} illustrates an overview of the proposed CARLS system. 
Model trainers are the main training jobs with an extended functionality to retrieve auxiliary information generated by the knowledge makers that are usually implemented by a fleet of servers or pipelines.
The model trainers and knowledge makers share the information via the knowledge bank, which stores the results generated by knowledge makers and enables efficient, asynchronous training.
Model trainers and knowledge makers may be deployed using different hardware platforms (\eg, CPUs, GPUs or TPUs).
Furthermore, model trainers and knowledge makers are communicating in an asynchronous fashion, meaning they operate independently without affecting the training speed.

The key observation to design CARLS for large-scale, efficient learning is that the compute can be parallelized at model-level (or logic-level) for many neural architectures; in other words, the interactions among multiple models are usually limited to a few ``connecting points'' such as the joint of each encoder output, or the aggregation of neighborhood embeddings in a graph neural network.
By employing the knowledge banks as the interface between individual modules, CARLS delegates the computation cost of the individual models from the model trainer to the knowledge makers. 
One potential issue of such an asynchronous mechanism is data freshness---some knowledge makers may generate results based on slightly outdated information.
In practice, we find the impacts of such an issue are controllable and not significant. 
In summary, the main advantages of the proposed CARLS are two-fold:
\begin{itemize}
    \item{\textbf{Scalability}: CARLS enables efficient operations to store and lookup intermediate knowledge for training complex models. For each training step, the intermediate knowledge can be fetched from a remote knowledge bank, rather than being recomputed across multiple training batches.}
    \item{\textbf{Flexibility}: CARLS allows constructing and updating the knowledge dynamically based on the current model state at each training step, as opposed to information preprocessed from a static input dataset or by a pretrained model.}
\end{itemize}

The proposed CARLS system is validated in a number of learning tasks which are considered to be very challenging or even infeasible before CARLS.
For example, training a graph neural network (GNN) stacked on top of a dense encoder (\eg, ResNet~\cite{resnet_2015} or Transformer~\cite{attention_all_you_need}), where each node features are encoded by the encoder and aggregated by a GNN, the training time becomes prohibitively expensive especially when the neighbor size grows large. 
With CARLS, we are able to train a graph-regularized model whose neighbor size is $10$ times larger than the largest one currently proposed by the literature~\cite{juan2020ultra}, without introducing any slowdown in the training speed.
We also demonstrate three learning paradigms as examples that can be scaled up by CARLS, including semi-supervised learning, curriculum learning, and multimodal learning. 

The remainder of this paper is organized as follows: after reviewing related work in Section~\ref{sec:related_work}, we elaborate the design of CARLS in Section~\ref{sec:overview}. In Section~\ref{sec:applications}, we demonstrate the learning paradigms that can be scaled up efficiently by CARLs. Finally, we conclude this paper in Section~\ref{sec:conclusion}.

\section{Related Work}
\label{sec:related_work}
Large-scale learning systems (\eg, models with more than 10G parameters) are often needed in language/image modeling, recommendation systems, 
or models with very large sparse feature set input (\eg, click through rate prediction models in digital advertising~\cite{zhou2018deep,zhao2019aibox}).
Directly relying on the design of existing deep learning libraries (\eg,~\cite{li2014scaling,abadi2016tensorflow}) is often limited in practice and 
challenges exist both in training and serving these large models.

Existing large language/image models are characterized by their sophisticated internal building blocks (\eg, the self-attention mechanism~\cite{attention_all_you_need} or the residual network~\cite{resnet_2015}) that are often stacked up to hundreds of layers (\eg,~\cite{bert,gpt1,gpt2,gpt3}).
Solely relying on the innovations of hardware (\eg, GPU or TPU) can hardly keep up with the explosion of model size.
To efficiently train these models, parallelisms/sparsity among the partitions of the model are often explored (\eg,~\cite{narayanan2019pipedream,lepikhin2020gshard}).
The mixture-of-experts (MoE) model is an example model with sparsity that allows different sub-models to be trained in parallel \cite{shazeer2017outrageously,narayanan2019pipedream}.
Most recently, the switch transformer~\cite{large_model_switch_transformers} is able to handle models with 1T parameters by solving a big MoE problem.
In order to serve these large models in small devices or more efficiently, the distillation technique~\cite{hinton2015distilling} or teacher-student networks are usually applied to compress the model.


Among industrial scale recommendation systems, the number of unique items (\eg, keywords, products) can easily reach up to billions or even trillions, which are usually represented as the softmax/logistic output layer of a model.
Since these output layers can be regarded as nearest neighbor search in an embedding space (one embedding for each item), a popular approach to efficient model training and inference is to compress the output embeddings via various hashing techniques~\cite{medini2019extreme,song2020large,zhang2018accelerated,tagami2017annexml}.
However, these compression/approximation approaches often come with the price of sacrificing modeling accuracy. 

There are also more recent interests in decoupling the memory part (\eg, sparse feature embeddings for both input and output layers) of a model from its structural part (\eg,~\cite{zhao2019aibox}),
therefore achieving efficient model training/inference, automatic model growth, as well as efficient multi-task learning.
For example, the DynamicEmbedding system~\cite{zeng2020dynamicembedding} employs a \emph{memory server} for serving the embedding data in a model with almost unlimited sparse feature set;
The Taskology system~\cite{lu2020taskology} employs an \emph{inference server} to facilitate multi-task training with consistency constraints.
In this work, we further extend this line of research by proposing a highly distributed system with both memory and inference components.

\section{Proposed CARLS}
\label{sec:overview}
In this section, we provide the details of the proposed CARLS framework, which consists of three major components that enable asynchronous representation learning across heterogeneous platforms:
\begin{itemize}
    \item{\textbf{Model Trainer}: responsible for both training various models and communicating with the knowledge bank to fetch augmented information to facilitate the training.}
    \item{\textbf{Knowledge Maker}: responsible for generating the knowledge requested by the model trainer. Specifically, a knowledge maker loads the latest checkpoint (saved by the model trainer) to make inference on auxiliary data, or discover new neighborhoods from examples with close representations learnt from the model.}
    \item{\textbf{Knowledge Bank}: responsible for storing and refreshing the knowledge generated by knowledge makers. A knowledge bank also serves the fetch requests from model trainers.}
\end{itemize}

\noindent Two key characteristics of CARLS are that (a) Model Trainers, Knowledge Banks, and Knowledge Makers can be operated asynchronously, and (b) these components can be deployed on heterogeneous hardware platforms (CPUs, GPUs, TPUs, etc). In the following sections, we describe the features that have been implemented, and leave a wide range of scenarios that are not yet implemented but accommodated by CARLS as a straight-line future work. 

\subsection{Knowledge Maker}\label{sec:03_km}
Knowledge makers play a critical role in extending the capacity of a learning system.
Conventionally, the model trainers process the data and compute all the information required for calculating losses.
In CARLS, knowledge makers are designed to dynamically calculate (or discover) the knowledge that can be exploited by model trainers for calculating the losses. Therefore, part of model trainers' workload can be shifted to knowledge makers, which helps the model trainers run faster or save capacity to accommodate larger models or training with more data.
Below we provide three example types of knowledge that are suitable to be processed or discovered by knowledge makers:
\begin{itemize}
    \item{\textbf{Graph structure and node embedding}: knowledge makers can load parts of the model---such as the node encoder of graph neural network---to update the knowledge. The graph structure can also be dynamically updated with the similarity between the computed node embeddings, as opposed to a given static graph (Section~\ref{sec:04_ssl}). 
    
    }    
    \item{\textbf{Augmented labels}: knowledge makers can load the whole model to (a) make inference on the missing labels for augmenting the training data, or (b) clean up noisy labels for providing a cleaner training dataset (Section~\ref{sec:04_curriculum}).
    }
    \item{\textbf{External Knowledge}: another advanced application of CARLS is to integrate the external knowledge/memeory for training~\cite{graves2016hybrid,weston2014memory}. In this case, the responsibility of knowledge makers is to pre-compute and process the related knowledge for enabling fast serving by the knowledge bank (Section~\ref{sec:04_multimodal})}
\end{itemize}

Knowledge makers keep the same machine states as model trainers by periodically loading the parameters from the latest checkpoints or parameter servers.
This way, the inferences made by the knowledge makers will be consistent with model trainers.
Knowledge makers can be implemented by a MapReduce~\cite{mapreduce} or Flume~\cite{chambers2010flumejava} pipeline, or by sending remote procedure calls (RPCs) to a remote inference server. 

\subsection{Knowledge Bank}\label{sec:03_kb}
The knowledge inferred by knowledge makers are stored in knowledge banks for fast serving model-training jobs.
Depending on the application type, a knowledge bank can be an external storage system---such as Bigtable~\cite{bigtable} or Spanner~\cite{spanner}---or a specialized embedding server~\cite{zeng2020dynamicembedding} responsible for updating internal parameters during back-propagation stage.
Some applications may require knowledge banks perform certain computations, such as retrieving top-N nearest neighbors given a query.
Note that the number of instances stored in knowledge banks can be very large (multi-millions to multi-billions).
To keep the computational latency constant---not growing as the data size grows, the knowledge banks are sharded and deployed in a distributed fashion~\cite{distributed_system_book}.

Below is a list of example applications enabled by knowledge banks:
\begin{itemize}
    \item{\textbf{Feature Lookup}: an instance's features (\eg, neighbor IDs from a graph, or labels) are stored as a protocol buffer~\cite{protobuffer} and keyed by the instance's unique ID. In this application, knowledge banks can be implemented directly by external storage systems without special treatments to store these feature information.}
    \item{\textbf{Embedding Lookup and Update}: in certain applications---\eg, node embeddings in the graph neural network or word embeddings in language modeling, the embeddings can be stored in a knowledge bank and will be continuously updated throughout the training process. To support back-propagate the gradients with respect to the embeddings, we leverage the DynamicEmbedding~\cite{zeng2020dynamicembedding} (as the implementation of knowledge banks) to update the embeddings with the gradient values.}
    \item{\textbf{Nearest Neighbors Lookup}: 
    efficient nearest neighbor search in the embedding space is important for solving large scale classification and retrieval tasks \cite{scann}. Unlike existing learning paradigms that either take the pre-computed nearest neighbors as input features or search over a very limited scope (\eg, within the same batch), CARLS enables searching over the embeddings kept in the knowledge bank, which is essentially the entire dataset---both labeled and unlabeled samples. Furthermore, using CARLS enables nearest-neighbor search over the dataset that grows or is updated during training. To facilitate efficient nearest neighbors search as the dataset grows, the computation is distributed into multiple shards and ScaNN~\cite{scann} can be applied for search space pruning and quantization.}
\end{itemize}

\paragraph{Lazy update for asynchronous gradient update}

While data consistency, isolation, and durability are delegated to the design of existing storage systems~\cite{bigtable,spanner} and computation efficiency and parallelism can be handled by existing server designs~\cite{zeng2020dynamicembedding}, special treatment is still required when multiple training jobs are attempting to update the gradients of the same embedding entry stored in knowledge bank.
In this case, simply guaranteeing atomicity may not be sufficient since this mechanism favors the last model that updates the gradients and ignores the contribution from other models.
CARLS resolves this issue by employing a lazy update scheme: caching the results of gradient update until the next lookup request arrives, or an expiration time is reached. The update is based on the average of all cached gradients with possible outlier detection.
With this lazy update mechanism, the overall training process is more stable compared with simple stochastic gradient descent.

\subsection{Model Trainer}\label{sec:03_trainer}
A model trainer is typically a machine-learning model equipped with a communication module to fetch the knowledge (of inferences) from knowledge banks. The communication module is usually implemented as customized operations or layers, enabling the funtionalities as listed in Section~\ref{sec:03_kb}. 
The communication module sends the request to the knowledge bank for retrieving additional information discovered during training.
The inputs of communication modules can come from the training data (\eg, sample IDs for fetching neighbor IDs), or the embedding values at a hidden layer (for retrieving the nearest neighbors).
Note that model trainers can be implemented with various platforms (\eg, TensorFlow or Pytorch) in a distributed fashion. Special cases need to be taken for the communication module when the trainers are running on TPU machines~\cite{jouppi2017datacenter_tpu}, which works best when the entire model fits into the on-chip memory.



\section{Learning Paradigms}
\label{sec:applications}
In the section, we provide three learning paradigms as examples that can be scaled up efficiently by CARLS, including but not limited to semi-supervised learning, curriculum learning, and multimodal learning.

\begin{figure}[t]
\centering
      \includegraphics[width=\linewidth]{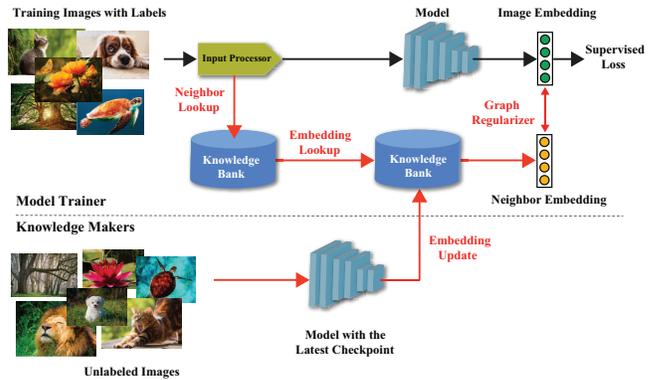}
  \caption{Training a graph-regularized model at scale with CARLS. Flow in red is added to enable graph regularization and required only during training. The model trainer and knowledge makers are running in parallel with a shared knowledge bank (KB). Knowledge makers keep loading the latest checkpoint generated by model trainers for making inferences and update the embeddings in KB. The input processor of the model trainer will look up training instances' neighborhood from the KB. After retrieving the neighbor info (e.g., neighbor IDs, edge weights), the neighbor IDs will be used to lookup neighbors' embeddings from the KB.}
 \label{fig:carls_ngm}
\end{figure}

\subsection{Semi-Supervised Learning}\label{sec:04_ssl}

Semi-supervised learning (SSL) is a machine learning paradigm that harnesses both labeled and unlabeled data to improve model prediction performance, and is particularly useful when labeled data is limited or expensive to obtain. However, despite the advantages and successes brought by SSL, using abundant unlabeled data also significantly increases the training computation. For example, graph-based methods are one type of popular SSL approaches, where each node's information (\eg, labels) can be iteratively updated by aggregating the information from neighboring nodes whether they are labeled or unlabeled~\cite{juan2020ultra, bui2018neural}. This means, in addition to computing or processing the labeled data, additional computation is required to process these neighboring nodes. In this case, CARLS deploy a fleet of knowledge makers to process these neighboring nodes, shifting the additional computation conventionally done by the model trainer to knowledge makers. Furthermore, CARLS eliminates redundant computations (\eg, one node being a common neighbor among several other nodes only needs to be processed once) by caching the results in the knowledge banks.

Figure~\ref{fig:carls_ngm} illustrates a graph-regularized model trained with CARLS. The training objective is to minimize the sum of the supervised loss and the graph regularizer, which is a pairwise distance between the embeddings of neighboring nodes. The neighbourhood can be constructed offline using existing signals (e.g., co-clicked image pairs ~\cite{juan2020ultra} or images in the same board~\cite{ying2018graph}), or created online during training (e.g., image augmentation~\cite{cubuk2019autoaugment}), and looked up from the knowledge bank in the trainer. 

If the neighbor embeddings are directly computed in the trainer, the training time and the trainer memory consumption will increase linearly in proportional to the number of neighbors~\cite{juan2020ultra}. To make the graph-regularized model scalable, the embeddings of each node/image are computed by knowledge makers and updated to the knowledge bank in parallel with the model trainer.

\begin{figure}[t]
\centering
      \includegraphics[width=\linewidth]{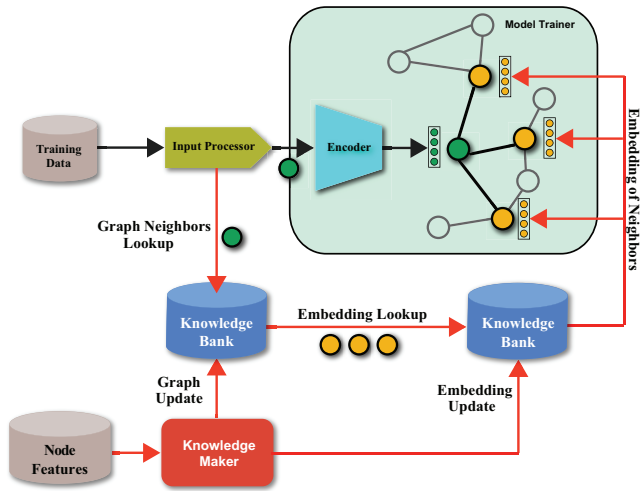}
  \caption{An illustration of the training pipeline of CARLS for a graph neural network stacked on top of a node encoder. The node encoder encodes the node features into embeddings and feed to the GNN. In the model trainer, each training input node can look up its sub-graph and the embeddings of each node in the subgraph from the KB. The knowledge makers can load the latest checkpoint to compute and update the node embeddings as well as the graph (based on the node embeddings) to the KB.}
 \label{fig:carls_encoder_gnn}
\end{figure}

Figure~\ref{fig:carls_encoder_gnn} illustrates another example of applying CARLS to enable efficient graph operations for large-scale graph learning. In this example, the model is a stack of an encoder (\eg ResNet or Transformer) and a graph neural network (GNN). It can be very challenging to train such a model, especially when the size of the subgraph is large, without the support of CARLS. 


\subsection{Curriculum Learning}\label{sec:04_curriculum}

\begin{figure}[t]
\centering
      \includegraphics[width=\linewidth]{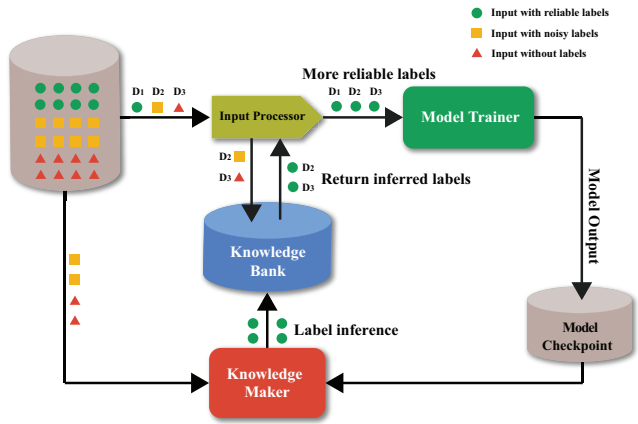}
  \caption{An illustration of the workflow of curriculum learning using CARLS. The labels of the input examples can be improved by the knowledge maker (\eg, using online label mining for unreliable labels or graph agreement model for inferring missing labels). }
 \label{fig:curriculum_learning}
\end{figure}

Curriculum learning~\cite{bengio2009curriculum} refers to a learning paradigm where the challenge of a learning task gradually increases over time for improving the model performance.
The increased challenge usually forces models to focus on learning additional information that is not well captured from the previous iterations~\cite{stretcu2019graph}.
This paradigm maps favorably to the CARLS framework where the additional information can be stored in the knowledge bank and the knowledge makers are responsible for generating this information required to increase the training complexity. 
Fig.~\ref{fig:curriculum_learning} illustrates the workflow of using CARLS for curriculum learning. Specifically, we demonstrate two examples: \emph{online label mining} deals with noisy labels, and the \emph{graph agreement model} deals with missing labels for some inputs. 

\subsubsection{Online label mining}
In large scale applications of image classification, the labels of each image are usually inferred from a rather noisy input (\eg, search queries or text captions associated to an image) that are often inconsistent with each other (\eg, wrong or missing labels). 
In such cases, multiple iterations are often needed to alternate between i) training the model based on noisy data and ii) refining the model based on the latest trained model. 
Such a process can be significantly accelerated if we can do i) and ii) in parallel.
In CARLS, i) can be achieved via a model trainer and ii) by knowledge makers, and the knowledge bank stores the newest inferred labeled for each instance.

\subsubsection{Graph Agreement Model}
Another example of curriculum learning is the graph agreement model for semi-supervised learning~\cite{stretcu2019graph}, where the labels of unknown training examples are inferred based on the nearest neighbors from labeled examples, where the nearest neighbors are calculated based on the embedding of the input (from the hidden layer of the trained model). 
In CARLS, this nearest neighbors search can be done by a knowledge maker and the inferred labels can be stored in the knowledge bank.

\subsection{Multimodal Learning}\label{sec:04_multimodal}
In addition to training a single deep model, CARLS can also be used for multimodal learning such as training an image-text two-tower model~\cite{jia2021scaling} and the modality nets~\cite{kaiser2017one} that has four models for language (text data), images, audio, and categorical data. 

\begin{figure}[t]
\centering
      \includegraphics[width=\linewidth]{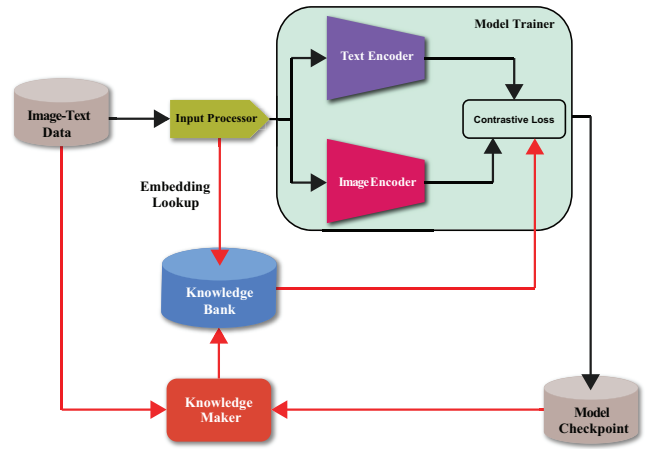}
  \caption{An illustration of the training pipeline of CARLS for a deep image-text two-tower model. Flow in red is added to enable lookup image/text embeddings and is required only during training. The distributed knowledge makers will compute the image/text embeddings and update the embeddings to the knowledge bank. The model trainer can look up the embeddings from the KB instead of computing the image/text embeddings through the image/text encoder.}
 \label{fig:carls_two_towers}
\end{figure}

Figure~\ref{fig:carls_two_towers} illustrates an example of a image-text deep two-tower model trained with CARLS, where the image and text encoders can be learned via a contrastive loss~\cite{chen2020simple} that pushes the embeddings of matched image-text pair together while pushing those of non-matched image-text pair apart. The image-text matching scores are computed as the cosine-similarity of the image embeddings and the text embeddings from the corresponding encoders. Although when the image/text encoder goes deeper it can often achieve better performance~\cite{jia2021scaling}, it is more challenging to train as the computational complexity grows significantly.  Instead of computing the image/text embeddings directly in the model trainer, we can utilize CARLS to lookup historical embeddings from the knowledge bank\footnote{In order to back-propagate the gradients back to the encoders, a subset of the examples in a training mini-batch should still be computed through the encoders in the model trainer.}. 

In addition to reducing the computation in the trainer, CARLS can also be used to improve the model performance. As shown in~\cite{jia2021scaling} that the model performance increases along with the number of random negatives used in the contrastive learning. As the embeddings of the random negatives can be looked up from the knowledge bank, we can easily scale up the number of random negatives. Besides, because each encoder can be considered as an independent model in the knowledge makers (by loading parts of the network from the checkpoint of the model), the image/text embeddings can be computed and updated to the knowledge bank independently. Thus, image/text data feed to the knowledge makers can go beyond the training image-text pairs.

\section{Conclusion}
\label{sec:conclusion}
In this paper, we proposed CARLS, a cross-platform, asynchronous learning framework that significantly extends the capacity of current deep learning frameworks by enabling dynamic knowledge sharing between model trainers and knowledge makers.
The proposed CARLS is particularly suitable for training large-scale neural models that benefit from additional knowledge inferred or discovered during training, such as graph neural networks and multi-modality models. 
CARLS also enables existing frameworks to dynamically learn with changes inferred or occurred during training, facilitating curriculum learning to be more efficient.
As a straight-line future work, we will extend CARLS to train more variety of learning models, design new APIs to streamline system setup and provide better support for heterogeneous hardware.


\balance
\bibliographystyle{ACM-Reference-Format}
\bibliography{acmart}


\begin{thebibliography}{55}


\ifx \showCODEN    \undefined \def \showCODEN     #1{\unskip}     \fi
\ifx \showDOI      \undefined \def \showDOI       #1{#1}\fi
\ifx \showISBNx    \undefined \def \showISBNx     #1{\unskip}     \fi
\ifx \showISBNxiii \undefined \def \showISBNxiii  #1{\unskip}     \fi
\ifx \showISSN     \undefined \def \showISSN      #1{\unskip}     \fi
\ifx \showLCCN     \undefined \def \showLCCN      #1{\unskip}     \fi
\ifx \shownote     \undefined \def \shownote      #1{#1}          \fi
\ifx \showarticletitle \undefined \def \showarticletitle #1{#1}   \fi
\ifx \showURL      \undefined \def \showURL       {\relax}        \fi
\providecommand\bibfield[2]{#2}
\providecommand\bibinfo[2]{#2}
\providecommand\natexlab[1]{#1}
\providecommand\showeprint[2][]{arXiv:#2}

\bibitem[\protect\citeauthoryear{??}{pro}{2016}]%
        {protobuffer}
 \bibinfo{year}{2016}\natexlab{}.
\newblock \bibinfo{title}{Protocol Buffers Language Guide}.
\newblock
\newblock


\bibitem[\protect\citeauthoryear{Abadi, Barham, Chen, Chen, Davis, Dean, Devin,
  Ghemawat, Irving, Isard, et~al\mbox{.}}{Abadi et~al\mbox{.}}{2016}]%
        {abadi2016tensorflow}
\bibfield{author}{\bibinfo{person}{Mart{\'\i}n Abadi}, \bibinfo{person}{Paul
  Barham}, \bibinfo{person}{Jianmin Chen}, \bibinfo{person}{Zhifeng Chen},
  \bibinfo{person}{Andy Davis}, \bibinfo{person}{Jeffrey Dean},
  \bibinfo{person}{Matthieu Devin}, \bibinfo{person}{Sanjay Ghemawat},
  \bibinfo{person}{Geoffrey Irving}, \bibinfo{person}{Michael Isard},
  {et~al\mbox{.}}} \bibinfo{year}{2016}\natexlab{}.
\newblock \showarticletitle{Tensorflow: a system for large-scale machine
  learning.}. In \bibinfo{booktitle}{\emph{USENIX Symposium on Operating
  Systems Design and Implementation (OSDI)}}, Vol.~\bibinfo{volume}{16}.
  \bibinfo{pages}{265--283}.
\newblock


\bibitem[\protect\citeauthoryear{Antol, Agrawal, Lu, Mitchell, Batra, Zitnick,
  and Parikh}{Antol et~al\mbox{.}}{2015}]%
        {antol2015vqa}
\bibfield{author}{\bibinfo{person}{Stanislaw Antol}, \bibinfo{person}{Aishwarya
  Agrawal}, \bibinfo{person}{Jiasen Lu}, \bibinfo{person}{Margaret Mitchell},
  \bibinfo{person}{Dhruv Batra}, \bibinfo{person}{C~Lawrence Zitnick}, {and}
  \bibinfo{person}{Devi Parikh}.} \bibinfo{year}{2015}\natexlab{}.
\newblock \showarticletitle{Vqa: Visual question answering}. In
  \bibinfo{booktitle}{\emph{Proceedings of the IEEE international conference on
  computer vision}}. \bibinfo{pages}{2425--2433}.
\newblock


\bibitem[\protect\citeauthoryear{Bengio, Louradour, Collobert, and
  Weston}{Bengio et~al\mbox{.}}{2009}]%
        {bengio2009curriculum}
\bibfield{author}{\bibinfo{person}{Yoshua Bengio},
  \bibinfo{person}{J{\'e}r{\^o}me Louradour}, \bibinfo{person}{Ronan
  Collobert}, {and} \bibinfo{person}{Jason Weston}.}
  \bibinfo{year}{2009}\natexlab{}.
\newblock \showarticletitle{Curriculum learning}. In
  \bibinfo{booktitle}{\emph{Proceedings of the 26th annual international
  conference on machine learning}}. \bibinfo{pages}{41--48}.
\newblock


\bibitem[\protect\citeauthoryear{Brown, Mann, Ryder, Subbiah, Kaplan, Dhariwal,
  Neelakantan, Shyam, Sastry, Askell, et~al\mbox{.}}{Brown
  et~al\mbox{.}}{2020}]%
        {gpt3}
\bibfield{author}{\bibinfo{person}{Tom~B Brown}, \bibinfo{person}{Benjamin
  Mann}, \bibinfo{person}{Nick Ryder}, \bibinfo{person}{Melanie Subbiah},
  \bibinfo{person}{Jared Kaplan}, \bibinfo{person}{Prafulla Dhariwal},
  \bibinfo{person}{Arvind Neelakantan}, \bibinfo{person}{Pranav Shyam},
  \bibinfo{person}{Girish Sastry}, \bibinfo{person}{Amanda Askell},
  {et~al\mbox{.}}} \bibinfo{year}{2020}\natexlab{}.
\newblock \showarticletitle{Language models are few-shot learners}.
\newblock \bibinfo{journal}{\emph{arXiv preprint arXiv:2005.14165}}
  (\bibinfo{year}{2020}).
\newblock


\bibitem[\protect\citeauthoryear{Bui, Ravi, and Ramavajjala}{Bui
  et~al\mbox{.}}{2018}]%
        {bui2018neural}
\bibfield{author}{\bibinfo{person}{Thang~D Bui}, \bibinfo{person}{Sujith Ravi},
  {and} \bibinfo{person}{Vivek Ramavajjala}.} \bibinfo{year}{2018}\natexlab{}.
\newblock \showarticletitle{Neural Graph Learning: Training Neural Networks
  Using Graphs}. In \bibinfo{booktitle}{\emph{ACM International Conference on
  Web Search and Data Mining}}.
\newblock


\bibitem[\protect\citeauthoryear{Chambers, Raniwala, Perry, Adams, Henry,
  Bradshaw, and Weizenbaum}{Chambers et~al\mbox{.}}{2010}]%
        {chambers2010flumejava}
\bibfield{author}{\bibinfo{person}{Craig Chambers}, \bibinfo{person}{Ashish
  Raniwala}, \bibinfo{person}{Frances Perry}, \bibinfo{person}{Stephen Adams},
  \bibinfo{person}{Robert~R Henry}, \bibinfo{person}{Robert Bradshaw}, {and}
  \bibinfo{person}{Nathan Weizenbaum}.} \bibinfo{year}{2010}\natexlab{}.
\newblock \showarticletitle{FlumeJava: easy, efficient data-parallel
  pipelines}.
\newblock \bibinfo{journal}{\emph{ACM Sigplan Notices}} \bibinfo{volume}{45},
  \bibinfo{number}{6} (\bibinfo{year}{2010}), \bibinfo{pages}{363--375}.
\newblock


\bibitem[\protect\citeauthoryear{Chang, Dean, Ghemawat, Hsieh, Wallach,
  Burrows, Chandra, Fikes, and Gruber}{Chang et~al\mbox{.}}{2006}]%
        {bigtable}
\bibfield{author}{\bibinfo{person}{Fay Chang}, \bibinfo{person}{Jeffrey Dean},
  \bibinfo{person}{Sanjay Ghemawat}, \bibinfo{person}{Wilson~C. Hsieh},
  \bibinfo{person}{Deborah~A. Wallach}, \bibinfo{person}{Mike Burrows},
  \bibinfo{person}{Tushar Chandra}, \bibinfo{person}{Andrew Fikes}, {and}
  \bibinfo{person}{Robert~E. Gruber}.} \bibinfo{year}{2006}\natexlab{}.
\newblock \showarticletitle{Bigtable: A Distributed Storage System for
  Structured Data}. In \bibinfo{booktitle}{\emph{7th {USENIX} Symposium on
  Operating Systems Design and Implementation (OSDI)}}.
  \bibinfo{pages}{205--218}.
\newblock


\bibitem[\protect\citeauthoryear{Chen, Kornblith, Norouzi, and Hinton}{Chen
  et~al\mbox{.}}{2020}]%
        {chen2020simple}
\bibfield{author}{\bibinfo{person}{Ting Chen}, \bibinfo{person}{Simon
  Kornblith}, \bibinfo{person}{Mohammad Norouzi}, {and}
  \bibinfo{person}{Geoffrey Hinton}.} \bibinfo{year}{2020}\natexlab{}.
\newblock \showarticletitle{A simple framework for contrastive learning of
  visual representations}. In \bibinfo{booktitle}{\emph{International
  conference on machine learning}}. PMLR, \bibinfo{pages}{1597--1607}.
\newblock


\bibitem[\protect\citeauthoryear{Collobert, Bengio, and
  Mari{\'e}thoz}{Collobert et~al\mbox{.}}{2002}]%
        {collobert2002torch}
\bibfield{author}{\bibinfo{person}{Ronan Collobert}, \bibinfo{person}{Samy
  Bengio}, {and} \bibinfo{person}{Johnny Mari{\'e}thoz}.}
  \bibinfo{year}{2002}\natexlab{}.
\newblock \bibinfo{booktitle}{\emph{Torch: a modular machine learning software
  library}}.
\newblock \bibinfo{type}{{T}echnical {R}eport}. \bibinfo{institution}{Idiap}.
\newblock


\bibitem[\protect\citeauthoryear{Corbett, Dean, Epstein, Fikes, Frost, Furman,
  Ghemawat, Gubarev, Heiser, Hochschild, Hsieh, Kanthak, Kogan, Li, Lloyd,
  Melnik, Mwaura, Nagle, Quinlan, Rao, Rolig, Woodford, Saito, Taylor,
  Szymaniak, and Wang}{Corbett et~al\mbox{.}}{2012}]%
        {spanner}
\bibfield{author}{\bibinfo{person}{James~C. Corbett}, \bibinfo{person}{Jeffrey
  Dean}, \bibinfo{person}{Michael Epstein}, \bibinfo{person}{Andrew Fikes},
  \bibinfo{person}{Christopher Frost}, \bibinfo{person}{JJ Furman},
  \bibinfo{person}{Sanjay Ghemawat}, \bibinfo{person}{Andrey Gubarev},
  \bibinfo{person}{Christopher Heiser}, \bibinfo{person}{Peter Hochschild},
  \bibinfo{person}{Wilson Hsieh}, \bibinfo{person}{Sebastian Kanthak},
  \bibinfo{person}{Eugene Kogan}, \bibinfo{person}{Hongyi Li},
  \bibinfo{person}{Alexander Lloyd}, \bibinfo{person}{Sergey Melnik},
  \bibinfo{person}{David Mwaura}, \bibinfo{person}{David Nagle},
  \bibinfo{person}{Sean Quinlan}, \bibinfo{person}{Rajesh Rao},
  \bibinfo{person}{Lindsay Rolig}, \bibinfo{person}{Dale Woodford},
  \bibinfo{person}{Yasushi Saito}, \bibinfo{person}{Christopher Taylor},
  \bibinfo{person}{Michal Szymaniak}, {and} \bibinfo{person}{Ruth Wang}.}
  \bibinfo{year}{2012}\natexlab{}.
\newblock \showarticletitle{Spanner: Google's Globally-Distributed Database}.
  In \bibinfo{booktitle}{\emph{OSDI}}.
\newblock


\bibitem[\protect\citeauthoryear{Cubuk, Zoph, Mane, Vasudevan, and Le}{Cubuk
  et~al\mbox{.}}{2019}]%
        {cubuk2019autoaugment}
\bibfield{author}{\bibinfo{person}{Ekin~D Cubuk}, \bibinfo{person}{Barret
  Zoph}, \bibinfo{person}{Dandelion Mane}, \bibinfo{person}{Vijay Vasudevan},
  {and} \bibinfo{person}{Quoc~V Le}.} \bibinfo{year}{2019}\natexlab{}.
\newblock \showarticletitle{Autoaugment: Learning augmentation strategies from
  data}. In \bibinfo{booktitle}{\emph{Proceedings of the IEEE/CVF Conference on
  Computer Vision and Pattern Recognition}}. \bibinfo{pages}{113--123}.
\newblock


\bibitem[\protect\citeauthoryear{Dean and Ghemawat}{Dean and Ghemawat}{2004}]%
        {mapreduce}
\bibfield{author}{\bibinfo{person}{Jeffrey Dean} {and} \bibinfo{person}{Sanjay
  Ghemawat}.} \bibinfo{year}{2004}\natexlab{}.
\newblock \showarticletitle{MapReduce: Simplified Data Processing on Large
  Clusters}. In \bibinfo{booktitle}{\emph{OSDI'04: Sixth Symposium on Operating
  System Design and Implementation}}. \bibinfo{address}{San Francisco, CA},
  \bibinfo{pages}{137--150}.
\newblock


\bibitem[\protect\citeauthoryear{Devlin, Chang, Lee, and Toutanova}{Devlin
  et~al\mbox{.}}{2018a}]%
        {bert}
\bibfield{author}{\bibinfo{person}{Jacob Devlin}, \bibinfo{person}{Ming{-}Wei
  Chang}, \bibinfo{person}{Kenton Lee}, {and} \bibinfo{person}{Kristina
  Toutanova}.} \bibinfo{year}{2018}\natexlab{a}.
\newblock \showarticletitle{{BERT:} Pre-training of Deep Bidirectional
  Transformers for Language Understanding}.
\newblock \bibinfo{journal}{\emph{CoRR}}  \bibinfo{volume}{abs/1810.04805}
  (\bibinfo{year}{2018}).
\newblock


\bibitem[\protect\citeauthoryear{Devlin, Chang, Lee, and Toutanova}{Devlin
  et~al\mbox{.}}{2018b}]%
        {devlin2018bert}
\bibfield{author}{\bibinfo{person}{Jacob Devlin}, \bibinfo{person}{Ming-Wei
  Chang}, \bibinfo{person}{Kenton Lee}, {and} \bibinfo{person}{Kristina
  Toutanova}.} \bibinfo{year}{2018}\natexlab{b}.
\newblock \showarticletitle{Bert: Pre-training of deep bidirectional
  transformers for language understanding}.
\newblock \bibinfo{journal}{\emph{arXiv preprint arXiv:1810.04805}}
  (\bibinfo{year}{2018}).
\newblock


\bibitem[\protect\citeauthoryear{Esteva, Robicquet, Ramsundar, Kuleshov,
  DePristo, Chou, Cui, Corrado, Thrun, and Dean}{Esteva et~al\mbox{.}}{2019}]%
        {deep_learning_in_health_care}
\bibfield{author}{\bibinfo{person}{Andre Esteva}, \bibinfo{person}{Alexandre
  Robicquet}, \bibinfo{person}{Bharath Ramsundar}, \bibinfo{person}{Volodymyr
  Kuleshov}, \bibinfo{person}{Mark DePristo}, \bibinfo{person}{Katherine Chou},
  \bibinfo{person}{Claire Cui}, \bibinfo{person}{Greg Corrado},
  \bibinfo{person}{Sebastian Thrun}, {and} \bibinfo{person}{Jeff Dean}.}
  \bibinfo{year}{2019}\natexlab{}.
\newblock \showarticletitle{A guide to deep learning in healthcare}.
\newblock \bibinfo{journal}{\emph{Nature medicine}} \bibinfo{volume}{25},
  \bibinfo{number}{1} (\bibinfo{year}{2019}), \bibinfo{pages}{24--29}.
\newblock


\bibitem[\protect\citeauthoryear{Fedus, Zoph, and Shazeer}{Fedus
  et~al\mbox{.}}{2021}]%
        {large_model_switch_transformers}
\bibfield{author}{\bibinfo{person}{William Fedus}, \bibinfo{person}{Barret
  Zoph}, {and} \bibinfo{person}{Noam Shazeer}.}
  \bibinfo{year}{2021}\natexlab{}.
\newblock \showarticletitle{Switch Transformers: Scaling to Trillion Parameter
  Models with Simple and Efficient Sparsity}.
\newblock \bibinfo{journal}{\emph{arXiv preprint arXiv:2101.03961}}
  (\bibinfo{year}{2021}).
\newblock


\bibitem[\protect\citeauthoryear{Goodfellow, Bengio, Courville, and
  Bengio}{Goodfellow et~al\mbox{.}}{2016}]%
        {goodfellow2016deep}
\bibfield{author}{\bibinfo{person}{Ian Goodfellow}, \bibinfo{person}{Yoshua
  Bengio}, \bibinfo{person}{Aaron Courville}, {and} \bibinfo{person}{Yoshua
  Bengio}.} \bibinfo{year}{2016}\natexlab{}.
\newblock \bibinfo{booktitle}{\emph{Deep learning}}. Vol.~\bibinfo{volume}{1}.
\newblock \bibinfo{publisher}{MIT press Cambridge}.
\newblock


\bibitem[\protect\citeauthoryear{Graves, Wayne, Reynolds, Harley, Danihelka,
  Grabska-Barwi{\'n}ska, Colmenarejo, Grefenstette, Ramalho, Agapiou,
  et~al\mbox{.}}{Graves et~al\mbox{.}}{2016}]%
        {graves2016hybrid}
\bibfield{author}{\bibinfo{person}{Alex Graves}, \bibinfo{person}{Greg Wayne},
  \bibinfo{person}{Malcolm Reynolds}, \bibinfo{person}{Tim Harley},
  \bibinfo{person}{Ivo Danihelka}, \bibinfo{person}{Agnieszka
  Grabska-Barwi{\'n}ska}, \bibinfo{person}{Sergio~G{\'o}mez Colmenarejo},
  \bibinfo{person}{Edward Grefenstette}, \bibinfo{person}{Tiago Ramalho},
  \bibinfo{person}{John Agapiou}, {et~al\mbox{.}}}
  \bibinfo{year}{2016}\natexlab{}.
\newblock \showarticletitle{Hybrid computing using a neural network with
  dynamic external memory}.
\newblock \bibinfo{journal}{\emph{Nature}} \bibinfo{volume}{538},
  \bibinfo{number}{7626} (\bibinfo{year}{2016}), \bibinfo{pages}{471--476}.
\newblock


\bibitem[\protect\citeauthoryear{Guo, Sun, Lindgren, Geng, Simcha, Chern, and
  Kumar}{Guo et~al\mbox{.}}{2020}]%
        {scann}
\bibfield{author}{\bibinfo{person}{Ruiqi Guo}, \bibinfo{person}{Philip Sun},
  \bibinfo{person}{Erik Lindgren}, \bibinfo{person}{Quan Geng},
  \bibinfo{person}{David Simcha}, \bibinfo{person}{Felix Chern}, {and}
  \bibinfo{person}{Sanjiv Kumar}.} \bibinfo{year}{2020}\natexlab{}.
\newblock \showarticletitle{Accelerating Large-Scale Inference with Anisotropic
  Vector Quantization}. In \bibinfo{booktitle}{\emph{International Conference
  on Machine Learning}}.
\newblock
\urldef\tempurl%
\url{https://arxiv.org/abs/1908.10396}
\showURL{%
\tempurl}


\bibitem[\protect\citeauthoryear{He, Zhang, Ren, and Sun}{He
  et~al\mbox{.}}{2015}]%
        {resnet_2015}
\bibfield{author}{\bibinfo{person}{Kaiming He}, \bibinfo{person}{Xiangyu
  Zhang}, \bibinfo{person}{Shaoqing Ren}, {and} \bibinfo{person}{Jian Sun}.}
  \bibinfo{year}{2015}\natexlab{}.
\newblock \showarticletitle{Deep Residual Learning for Image Recognition}. In
  \bibinfo{booktitle}{\emph{Computer Vision and Pattern Recognition (CVPR)}}.
\newblock


\bibitem[\protect\citeauthoryear{Hinton, Vinyals, and Dean}{Hinton
  et~al\mbox{.}}{2015}]%
        {hinton2015distilling}
\bibfield{author}{\bibinfo{person}{Geoffrey Hinton}, \bibinfo{person}{Oriol
  Vinyals}, {and} \bibinfo{person}{Jeff Dean}.}
  \bibinfo{year}{2015}\natexlab{}.
\newblock \showarticletitle{Distilling the knowledge in a neural network}.
\newblock \bibinfo{journal}{\emph{arXiv preprint arXiv:1503.02531}}
  (\bibinfo{year}{2015}).
\newblock


\bibitem[\protect\citeauthoryear{Jia, Yang, Xia, Chen, Parekh, Pham, Le, Sung,
  Li, and Duerig}{Jia et~al\mbox{.}}{2021}]%
        {jia2021scaling}
\bibfield{author}{\bibinfo{person}{Chao Jia}, \bibinfo{person}{Yinfei Yang},
  \bibinfo{person}{Ye Xia}, \bibinfo{person}{Yi-Ting Chen},
  \bibinfo{person}{Zarana Parekh}, \bibinfo{person}{Hieu Pham},
  \bibinfo{person}{Quoc~V. Le}, \bibinfo{person}{Yunhsuan Sung},
  \bibinfo{person}{Zhen Li}, {and} \bibinfo{person}{Tom Duerig}.}
  \bibinfo{year}{2021}\natexlab{}.
\newblock \bibinfo{title}{Scaling Up Visual and Vision-Language Representation
  Learning With Noisy Text Supervision}.
\newblock
\newblock
\showeprint[arxiv]{cs.CV/2102.05918}


\bibitem[\protect\citeauthoryear{Jouppi, Young, Patil, Patterson, Agrawal,
  Bajwa, Bates, Bhatia, Boden, Borchers, et~al\mbox{.}}{Jouppi
  et~al\mbox{.}}{2017}]%
        {jouppi2017datacenter_tpu}
\bibfield{author}{\bibinfo{person}{Norman~P Jouppi}, \bibinfo{person}{Cliff
  Young}, \bibinfo{person}{Nishant Patil}, \bibinfo{person}{David Patterson},
  \bibinfo{person}{Gaurav Agrawal}, \bibinfo{person}{Raminder Bajwa},
  \bibinfo{person}{Sarah Bates}, \bibinfo{person}{Suresh Bhatia},
  \bibinfo{person}{Nan Boden}, \bibinfo{person}{Al Borchers}, {et~al\mbox{.}}}
  \bibinfo{year}{2017}\natexlab{}.
\newblock \showarticletitle{In-datacenter performance analysis of a tensor
  processing unit}. In \bibinfo{booktitle}{\emph{Proceedings of the 44th annual
  international symposium on computer architecture}}. \bibinfo{pages}{1--12}.
\newblock


\bibitem[\protect\citeauthoryear{Juan, Lu, Li, Peng, Timofeev, Chen, Gao,
  Duerig, Tomkins, and Ravi}{Juan et~al\mbox{.}}{2020}]%
        {juan2020ultra}
\bibfield{author}{\bibinfo{person}{Da-Cheng Juan}, \bibinfo{person}{Chun-Ta
  Lu}, \bibinfo{person}{Zhen Li}, \bibinfo{person}{Futang Peng},
  \bibinfo{person}{Aleksei Timofeev}, \bibinfo{person}{Yi-Ting Chen},
  \bibinfo{person}{Yaxi Gao}, \bibinfo{person}{Tom Duerig},
  \bibinfo{person}{Andrew Tomkins}, {and} \bibinfo{person}{Sujith Ravi}.}
  \bibinfo{year}{2020}\natexlab{}.
\newblock \showarticletitle{Ultra Fine-Grained Image Semantic Embedding}. In
  \bibinfo{booktitle}{\emph{Proceedings of the 13th International Conference on
  Web Search and Data Mining}}. \bibinfo{pages}{277--285}.
\newblock


\bibitem[\protect\citeauthoryear{Kaiser, Gomez, Shazeer, Vaswani, Parmar,
  Jones, and Uszkoreit}{Kaiser et~al\mbox{.}}{2017}]%
        {kaiser2017one}
\bibfield{author}{\bibinfo{person}{Lukasz Kaiser}, \bibinfo{person}{Aidan~N
  Gomez}, \bibinfo{person}{Noam Shazeer}, \bibinfo{person}{Ashish Vaswani},
  \bibinfo{person}{Niki Parmar}, \bibinfo{person}{Llion Jones}, {and}
  \bibinfo{person}{Jakob Uszkoreit}.} \bibinfo{year}{2017}\natexlab{}.
\newblock \showarticletitle{One model to learn them all}.
\newblock \bibinfo{journal}{\emph{arXiv preprint arXiv:1706.05137}}
  (\bibinfo{year}{2017}).
\newblock


\bibitem[\protect\citeauthoryear{Kandel, Schwartz, Jessell, of~Biochemistry,
  Jessell, Siegelbaum, and Hudspeth}{Kandel et~al\mbox{.}}{2000}]%
        {kandel2000principles}
\bibfield{author}{\bibinfo{person}{Eric~R Kandel}, \bibinfo{person}{James~H
  Schwartz}, \bibinfo{person}{Thomas~M Jessell}, \bibinfo{person}{Department of
  Biochemistry}, \bibinfo{person}{Molecular Biophysics~Thomas Jessell},
  \bibinfo{person}{Steven Siegelbaum}, {and} \bibinfo{person}{AJ Hudspeth}.}
  \bibinfo{year}{2000}\natexlab{}.
\newblock \bibinfo{booktitle}{\emph{Principles of neural science}}.
  Vol.~\bibinfo{volume}{4}.
\newblock \bibinfo{publisher}{McGraw-hill New York}.
\newblock


\bibitem[\protect\citeauthoryear{Kipf and Welling}{Kipf and Welling}{2016}]%
        {kipf2016semi}
\bibfield{author}{\bibinfo{person}{Thomas~N Kipf} {and} \bibinfo{person}{Max
  Welling}.} \bibinfo{year}{2016}\natexlab{}.
\newblock \showarticletitle{Semi-supervised classification with graph
  convolutional networks}.
\newblock \bibinfo{journal}{\emph{arXiv preprint arXiv:1609.02907}}
  (\bibinfo{year}{2016}).
\newblock


\bibitem[\protect\citeauthoryear{LeCun, Bengio, and Hinton}{LeCun
  et~al\mbox{.}}{2015}]%
        {deep_learning_nature}
\bibfield{author}{\bibinfo{person}{Yann LeCun}, \bibinfo{person}{Yoshua
  Bengio}, {and} \bibinfo{person}{Geoffrey Hinton}.}
  \bibinfo{year}{2015}\natexlab{}.
\newblock \showarticletitle{Deep Learning}.
\newblock \bibinfo{journal}{\emph{Nature}}  \bibinfo{volume}{521}
  (\bibinfo{year}{2015}), \bibinfo{pages}{436--444}.
\newblock


\bibitem[\protect\citeauthoryear{Lepikhin, Lee, Xu, Chen, Firat, Huang, Krikun,
  Shazeer, and Chen}{Lepikhin et~al\mbox{.}}{2020}]%
        {lepikhin2020gshard}
\bibfield{author}{\bibinfo{person}{Dmitry Lepikhin},
  \bibinfo{person}{HyoukJoong Lee}, \bibinfo{person}{Yuanzhong Xu},
  \bibinfo{person}{Dehao Chen}, \bibinfo{person}{Orhan Firat},
  \bibinfo{person}{Yanping Huang}, \bibinfo{person}{Maxim Krikun},
  \bibinfo{person}{Noam Shazeer}, {and} \bibinfo{person}{Zhifeng Chen}.}
  \bibinfo{year}{2020}\natexlab{}.
\newblock \showarticletitle{Gshard: Scaling giant models with conditional
  computation and automatic sharding}.
\newblock \bibinfo{journal}{\emph{arXiv preprint arXiv:2006.16668}}
  (\bibinfo{year}{2020}).
\newblock


\bibitem[\protect\citeauthoryear{Li, Andersen, Park, Smola, Ahmed, Josifovski,
  Long, Shekita, and Su}{Li et~al\mbox{.}}{2014}]%
        {li2014scaling}
\bibfield{author}{\bibinfo{person}{Mu Li}, \bibinfo{person}{David~G Andersen},
  \bibinfo{person}{Jun~Woo Park}, \bibinfo{person}{Alexander~J Smola},
  \bibinfo{person}{Amr Ahmed}, \bibinfo{person}{Vanja Josifovski},
  \bibinfo{person}{James Long}, \bibinfo{person}{Eugene~J Shekita}, {and}
  \bibinfo{person}{Bor-Yiing Su}.} \bibinfo{year}{2014}\natexlab{}.
\newblock \showarticletitle{Scaling distributed machine learning with the
  parameter server}. In \bibinfo{booktitle}{\emph{11th $\{$USENIX$\}$ Symposium
  on Operating Systems Design and Implementation ($\{$OSDI$\}$ 14)}}.
  \bibinfo{pages}{583--598}.
\newblock


\bibitem[\protect\citeauthoryear{Lu, Goswami, Rohrbach, Parikh, and Lee}{Lu
  et~al\mbox{.}}{2020a}]%
        {lu202012}
\bibfield{author}{\bibinfo{person}{Jiasen Lu}, \bibinfo{person}{Vedanuj
  Goswami}, \bibinfo{person}{Marcus Rohrbach}, \bibinfo{person}{Devi Parikh},
  {and} \bibinfo{person}{Stefan Lee}.} \bibinfo{year}{2020}\natexlab{a}.
\newblock \showarticletitle{12-in-1: Multi-task vision and language
  representation learning}. In \bibinfo{booktitle}{\emph{Proceedings of the
  IEEE/CVF Conference on Computer Vision and Pattern Recognition}}.
  \bibinfo{pages}{10437--10446}.
\newblock


\bibitem[\protect\citeauthoryear{Lu, Pirk, Dlabal, Brohan, Pasad, Chen, Casser,
  Angelova, and Gordon}{Lu et~al\mbox{.}}{2020b}]%
        {lu2020taskology}
\bibfield{author}{\bibinfo{person}{Yao Lu}, \bibinfo{person}{S{\"o}ren Pirk},
  \bibinfo{person}{Jan Dlabal}, \bibinfo{person}{Anthony Brohan},
  \bibinfo{person}{Ankita Pasad}, \bibinfo{person}{Zhao Chen},
  \bibinfo{person}{Vincent Casser}, \bibinfo{person}{Anelia Angelova}, {and}
  \bibinfo{person}{Ariel Gordon}.} \bibinfo{year}{2020}\natexlab{b}.
\newblock \showarticletitle{Taskology: Utilizing Task Relations at Scale}.
\newblock \bibinfo{journal}{\emph{arXiv preprint arXiv:2005.07289}}
  (\bibinfo{year}{2020}).
\newblock


\bibitem[\protect\citeauthoryear{Medini, Huang, Wang, Mohan, and
  Shrivastava}{Medini et~al\mbox{.}}{2019}]%
        {medini2019extreme}
\bibfield{author}{\bibinfo{person}{Tharun Medini}, \bibinfo{person}{Qixuan
  Huang}, \bibinfo{person}{Yiqiu Wang}, \bibinfo{person}{Vijai Mohan}, {and}
  \bibinfo{person}{Anshumali Shrivastava}.} \bibinfo{year}{2019}\natexlab{}.
\newblock \showarticletitle{Extreme classification in log memory using
  count-min sketch: A case study of amazon search with 50m products}.
\newblock \bibinfo{journal}{\emph{arXiv preprint arXiv:1910.13830}}
  (\bibinfo{year}{2019}).
\newblock


\bibitem[\protect\citeauthoryear{Narayanan, Harlap, Phanishayee, Seshadri,
  Devanur, Ganger, Gibbons, and Zaharia}{Narayanan et~al\mbox{.}}{2019}]%
        {narayanan2019pipedream}
\bibfield{author}{\bibinfo{person}{Deepak Narayanan}, \bibinfo{person}{Aaron
  Harlap}, \bibinfo{person}{Amar Phanishayee}, \bibinfo{person}{Vivek
  Seshadri}, \bibinfo{person}{Nikhil~R Devanur}, \bibinfo{person}{Gregory~R
  Ganger}, \bibinfo{person}{Phillip~B Gibbons}, {and} \bibinfo{person}{Matei
  Zaharia}.} \bibinfo{year}{2019}\natexlab{}.
\newblock \showarticletitle{PipeDream: generalized pipeline parallelism for DNN
  training}. In \bibinfo{booktitle}{\emph{Proceedings of the 27th ACM Symposium
  on Operating Systems Principles}}. \bibinfo{pages}{1--15}.
\newblock


\bibitem[\protect\citeauthoryear{Py{T}orch}{Py{T}orch}{2018}]%
        {pytorch}
\bibfield{author}{\bibinfo{person}{Py{T}orch}.}
  \bibinfo{year}{2018}\natexlab{}.
\newblock
\newblock
\urldef\tempurl%
\url{http://pytorch.org}
\showURL{%
\tempurl}
\newblock
\shownote{http://pytorch.org.}


\bibitem[\protect\citeauthoryear{Radford, Kim, Hallacy, Ramesh, Goh, Agarwal,
  Sastry, Askell, Mishkin, Clark, et~al\mbox{.}}{Radford et~al\mbox{.}}{[n.
  d.]}]%
        {radford2learning}
\bibfield{author}{\bibinfo{person}{Alec Radford}, \bibinfo{person}{Jong~Wook
  Kim}, \bibinfo{person}{Chris Hallacy}, \bibinfo{person}{Aditya Ramesh},
  \bibinfo{person}{Gabriel Goh}, \bibinfo{person}{Sandhini Agarwal},
  \bibinfo{person}{Girish Sastry}, \bibinfo{person}{Amanda Askell},
  \bibinfo{person}{Pamela Mishkin}, \bibinfo{person}{Jack Clark},
  {et~al\mbox{.}}} \bibinfo{year}{[n. d.]}\natexlab{}.
\newblock \showarticletitle{Learning Transferable Visual Models From Natural
  Language Supervision}.
\newblock \bibinfo{journal}{\emph{Image}}  \bibinfo{volume}{2}
  (\bibinfo{year}{[n. d.]}), \bibinfo{pages}{T2}.
\newblock


\bibitem[\protect\citeauthoryear{Radford, Narasimhan, Salimans, and
  Sutskever}{Radford et~al\mbox{.}}{2018}]%
        {gpt1}
\bibfield{author}{\bibinfo{person}{Alec Radford}, \bibinfo{person}{Karthik
  Narasimhan}, \bibinfo{person}{Tim Salimans}, {and} \bibinfo{person}{Ilya
  Sutskever}.} \bibinfo{year}{2018}\natexlab{}.
\newblock \showarticletitle{Improving language understanding by generative
  pre-training}.
\newblock  (\bibinfo{year}{2018}).
\newblock


\bibitem[\protect\citeauthoryear{Radford, Wu, Child, Luan, Amodei, and
  Sutskever}{Radford et~al\mbox{.}}{2019}]%
        {gpt2}
\bibfield{author}{\bibinfo{person}{Alec Radford}, \bibinfo{person}{Jeffrey Wu},
  \bibinfo{person}{Rewon Child}, \bibinfo{person}{David Luan},
  \bibinfo{person}{Dario Amodei}, {and} \bibinfo{person}{Ilya Sutskever}.}
  \bibinfo{year}{2019}\natexlab{}.
\newblock \showarticletitle{Language models are unsupervised multitask
  learners}.
\newblock \bibinfo{journal}{\emph{OpenAI blog}} \bibinfo{volume}{1},
  \bibinfo{number}{8} (\bibinfo{year}{2019}), \bibinfo{pages}{9}.
\newblock


\bibitem[\protect\citeauthoryear{Raffel, Shazeer, Roberts, Lee, Narang, Matena,
  Zhou, Li, and Liu}{Raffel et~al\mbox{.}}{2020}]%
        {2020t5}
\bibfield{author}{\bibinfo{person}{Colin Raffel}, \bibinfo{person}{Noam
  Shazeer}, \bibinfo{person}{Adam Roberts}, \bibinfo{person}{Katherine Lee},
  \bibinfo{person}{Sharan Narang}, \bibinfo{person}{Michael Matena},
  \bibinfo{person}{Yanqi Zhou}, \bibinfo{person}{Wei Li}, {and}
  \bibinfo{person}{Peter~J. Liu}.} \bibinfo{year}{2020}\natexlab{}.
\newblock \showarticletitle{Exploring the Limits of Transfer Learning with a
  Unified Text-to-Text Transformer}.
\newblock \bibinfo{journal}{\emph{Journal of Machine Learning Research}}
  \bibinfo{volume}{21}, \bibinfo{number}{140} (\bibinfo{year}{2020}),
  \bibinfo{pages}{1--67}.
\newblock
\urldef\tempurl%
\url{http://jmlr.org/papers/v21/20-074.html}
\showURL{%
\tempurl}


\bibitem[\protect\citeauthoryear{Shazeer, Mirhoseini, Maziarz, Davis, Le,
  Hinton, and Dean}{Shazeer et~al\mbox{.}}{2017}]%
        {shazeer2017outrageously}
\bibfield{author}{\bibinfo{person}{Noam Shazeer}, \bibinfo{person}{Azalia
  Mirhoseini}, \bibinfo{person}{Krzysztof Maziarz}, \bibinfo{person}{Andy
  Davis}, \bibinfo{person}{Quoc Le}, \bibinfo{person}{Geoffrey Hinton}, {and}
  \bibinfo{person}{Jeff Dean}.} \bibinfo{year}{2017}\natexlab{}.
\newblock \showarticletitle{Outrageously large neural networks: The
  sparsely-gated mixture-of-experts layer}.
\newblock \bibinfo{journal}{\emph{arXiv preprint arXiv:1701.06538}}
  (\bibinfo{year}{2017}).
\newblock


\bibitem[\protect\citeauthoryear{Song, Pan, Zhao, Yang, Chen, Zhang, Xu, and
  Jin}{Song et~al\mbox{.}}{2020}]%
        {song2020large}
\bibfield{author}{\bibinfo{person}{Liuyihan Song}, \bibinfo{person}{Pan Pan},
  \bibinfo{person}{Kang Zhao}, \bibinfo{person}{Hao Yang},
  \bibinfo{person}{Yiming Chen}, \bibinfo{person}{Yingya Zhang},
  \bibinfo{person}{Yinghui Xu}, {and} \bibinfo{person}{Rong Jin}.}
  \bibinfo{year}{2020}\natexlab{}.
\newblock \showarticletitle{Large-Scale Training System for 100-Million
  Classification at Alibaba}. In \bibinfo{booktitle}{\emph{Proceedings of the
  26th ACM SIGKDD International Conference on Knowledge Discovery \& Data
  Mining}}. \bibinfo{pages}{2909--2930}.
\newblock


\bibitem[\protect\citeauthoryear{Stretcu, Viswanathan, Movshovitz-Attias,
  Platanios, Ravi, and Tomkins}{Stretcu et~al\mbox{.}}{2019}]%
        {stretcu2019graph}
\bibfield{author}{\bibinfo{person}{Otilia Stretcu},
  \bibinfo{person}{Krishnamurthy Viswanathan}, \bibinfo{person}{Dana
  Movshovitz-Attias}, \bibinfo{person}{Anthony Platanios},
  \bibinfo{person}{Sujith Ravi}, {and} \bibinfo{person}{Andrew Tomkins}.}
  \bibinfo{year}{2019}\natexlab{}.
\newblock \showarticletitle{Graph agreement models for semi-supervised
  learning}.
\newblock  (\bibinfo{year}{2019}).
\newblock


\bibitem[\protect\citeauthoryear{Tagami}{Tagami}{2017}]%
        {tagami2017annexml}
\bibfield{author}{\bibinfo{person}{Yukihiro Tagami}.}
  \bibinfo{year}{2017}\natexlab{}.
\newblock \showarticletitle{Annexml: Approximate nearest neighbor search for
  extreme multi-label classification}. In \bibinfo{booktitle}{\emph{Proceedings
  of the 23rd ACM SIGKDD international conference on knowledge discovery and
  data mining}}. \bibinfo{pages}{455--464}.
\newblock


\bibitem[\protect\citeauthoryear{van Steen and Tanenbaum}{van Steen and
  Tanenbaum}{2017}]%
        {distributed_system_book}
\bibfield{author}{\bibinfo{person}{Maarten van Steen} {and}
  \bibinfo{person}{Andrew~S. Tanenbaum}.} \bibinfo{year}{2017}\natexlab{}.
\newblock \bibinfo{booktitle}{\emph{Distributed Systems}}.
\newblock \bibinfo{publisher}{CreateSpace Independent Publishing Platform}.
\newblock


\bibitem[\protect\citeauthoryear{Vaswani, Shazeer, Parmar, Uszkoreit, Jones,
  Gomez, Kaiser, and Polosukhin}{Vaswani et~al\mbox{.}}{2017}]%
        {attention_all_you_need}
\bibfield{author}{\bibinfo{person}{Ashish Vaswani}, \bibinfo{person}{Noam
  Shazeer}, \bibinfo{person}{Niki Parmar}, \bibinfo{person}{Jakob Uszkoreit},
  \bibinfo{person}{Llion Jones}, \bibinfo{person}{Aidan~N Gomez},
  \bibinfo{person}{Lukasz Kaiser}, {and} \bibinfo{person}{Illia Polosukhin}.}
  \bibinfo{year}{2017}\natexlab{}.
\newblock \showarticletitle{Attention is All you Need}.
\newblock In \bibinfo{booktitle}{\emph{Advances in Neural Information
  Processing Systems (NIPS)}}. \bibinfo{pages}{5998--6008}.
\newblock


\bibitem[\protect\citeauthoryear{Veli{\v{c}}kovi{\'c}, Cucurull, Casanova,
  Romero, Lio, and Bengio}{Veli{\v{c}}kovi{\'c} et~al\mbox{.}}{2017}]%
        {velivckovic2017graph}
\bibfield{author}{\bibinfo{person}{Petar Veli{\v{c}}kovi{\'c}},
  \bibinfo{person}{Guillem Cucurull}, \bibinfo{person}{Arantxa Casanova},
  \bibinfo{person}{Adriana Romero}, \bibinfo{person}{Pietro Lio}, {and}
  \bibinfo{person}{Yoshua Bengio}.} \bibinfo{year}{2017}\natexlab{}.
\newblock \showarticletitle{Graph attention networks}.
\newblock \bibinfo{journal}{\emph{arXiv preprint arXiv:1710.10903}}
  (\bibinfo{year}{2017}).
\newblock


\bibitem[\protect\citeauthoryear{Weston, Chopra, and Bordes}{Weston
  et~al\mbox{.}}{2014}]%
        {weston2014memory}
\bibfield{author}{\bibinfo{person}{Jason Weston}, \bibinfo{person}{Sumit
  Chopra}, {and} \bibinfo{person}{Antoine Bordes}.}
  \bibinfo{year}{2014}\natexlab{}.
\newblock \showarticletitle{Memory networks}.
\newblock \bibinfo{journal}{\emph{arXiv preprint arXiv:1410.3916}}
  (\bibinfo{year}{2014}).
\newblock


\bibitem[\protect\citeauthoryear{Ying, He, Chen, Eksombatchai, Hamilton, and
  Leskovec}{Ying et~al\mbox{.}}{2018}]%
        {ying2018graph}
\bibfield{author}{\bibinfo{person}{Rex Ying}, \bibinfo{person}{Ruining He},
  \bibinfo{person}{Kaifeng Chen}, \bibinfo{person}{Pong Eksombatchai},
  \bibinfo{person}{William~L Hamilton}, {and} \bibinfo{person}{Jure Leskovec}.}
  \bibinfo{year}{2018}\natexlab{}.
\newblock \showarticletitle{Graph convolutional neural networks for web-scale
  recommender systems}. In \bibinfo{booktitle}{\emph{Proceedings of the 24th
  ACM SIGKDD International Conference on Knowledge Discovery \& Data Mining}}.
  \bibinfo{pages}{974--983}.
\newblock


\bibitem[\protect\citeauthoryear{Young, Lai, Hodosh, and Hockenmaier}{Young
  et~al\mbox{.}}{2014}]%
        {young2014image}
\bibfield{author}{\bibinfo{person}{Peter Young}, \bibinfo{person}{Alice Lai},
  \bibinfo{person}{Micah Hodosh}, {and} \bibinfo{person}{Julia Hockenmaier}.}
  \bibinfo{year}{2014}\natexlab{}.
\newblock \showarticletitle{From image descriptions to visual denotations: New
  similarity metrics for semantic inference over event descriptions}.
\newblock \bibinfo{journal}{\emph{Transactions of the Association for
  Computational Linguistics}}  \bibinfo{volume}{2} (\bibinfo{year}{2014}),
  \bibinfo{pages}{67--78}.
\newblock


\bibitem[\protect\citeauthoryear{Zellers, Bisk, Farhadi, and Choi}{Zellers
  et~al\mbox{.}}{2019}]%
        {zellers2019recognition}
\bibfield{author}{\bibinfo{person}{Rowan Zellers}, \bibinfo{person}{Yonatan
  Bisk}, \bibinfo{person}{Ali Farhadi}, {and} \bibinfo{person}{Yejin Choi}.}
  \bibinfo{year}{2019}\natexlab{}.
\newblock \showarticletitle{From recognition to cognition: Visual commonsense
  reasoning}. In \bibinfo{booktitle}{\emph{Proceedings of the IEEE/CVF
  Conference on Computer Vision and Pattern Recognition}}.
  \bibinfo{pages}{6720--6731}.
\newblock


\bibitem[\protect\citeauthoryear{Zeng, Zuo, and Shen}{Zeng
  et~al\mbox{.}}{2020}]%
        {zeng2020dynamicembedding}
\bibfield{author}{\bibinfo{person}{Yun Zeng}, \bibinfo{person}{Siqi Zuo}, {and}
  \bibinfo{person}{Dongcai Shen}.} \bibinfo{year}{2020}\natexlab{}.
\newblock \showarticletitle{DynamicEmbedding: Extending TensorFlow for
  Colossal-Scale Applications}.
\newblock \bibinfo{journal}{\emph{arXiv preprint arXiv:2004.08366}}
  (\bibinfo{year}{2020}).
\newblock


\bibitem[\protect\citeauthoryear{Zhang, Yang, Yan, and Lin}{Zhang
  et~al\mbox{.}}{2018}]%
        {zhang2018accelerated}
\bibfield{author}{\bibinfo{person}{Xingcheng Zhang}, \bibinfo{person}{Lei
  Yang}, \bibinfo{person}{Junjie Yan}, {and} \bibinfo{person}{Dahua Lin}.}
  \bibinfo{year}{2018}\natexlab{}.
\newblock \showarticletitle{Accelerated training for massive classification via
  dynamic class selection}. In \bibinfo{booktitle}{\emph{Proceedings of the
  AAAI Conference on Artificial Intelligence}}, Vol.~\bibinfo{volume}{32}.
\newblock


\bibitem[\protect\citeauthoryear{Zhao, Zhang, Xie, Qian, Jia, and Li}{Zhao
  et~al\mbox{.}}{2019}]%
        {zhao2019aibox}
\bibfield{author}{\bibinfo{person}{Weijie Zhao}, \bibinfo{person}{Jingyuan
  Zhang}, \bibinfo{person}{Deping Xie}, \bibinfo{person}{Yulei Qian},
  \bibinfo{person}{Ronglai Jia}, {and} \bibinfo{person}{Ping Li}.}
  \bibinfo{year}{2019}\natexlab{}.
\newblock \showarticletitle{AIBox: CTR prediction model training on a single
  node}. In \bibinfo{booktitle}{\emph{Proceedings of the 28th ACM International
  Conference on Information and Knowledge Management}}.
  \bibinfo{pages}{319--328}.
\newblock


\bibitem[\protect\citeauthoryear{Zhou, Zhu, Song, Fan, Zhu, Ma, Yan, Jin, Li,
  and Gai}{Zhou et~al\mbox{.}}{2018}]%
        {zhou2018deep}
\bibfield{author}{\bibinfo{person}{Guorui Zhou}, \bibinfo{person}{Xiaoqiang
  Zhu}, \bibinfo{person}{Chenru Song}, \bibinfo{person}{Ying Fan},
  \bibinfo{person}{Han Zhu}, \bibinfo{person}{Xiao Ma},
  \bibinfo{person}{Yanghui Yan}, \bibinfo{person}{Junqi Jin},
  \bibinfo{person}{Han Li}, {and} \bibinfo{person}{Kun Gai}.}
  \bibinfo{year}{2018}\natexlab{}.
\newblock \showarticletitle{Deep interest network for click-through rate
  prediction}. In \bibinfo{booktitle}{\emph{Proceedings of the 24th ACM SIGKDD
  International Conference on Knowledge Discovery \& Data Mining}}.
  \bibinfo{pages}{1059--1068}.
\newblock


\end{thebibliography}

\end{document}